\pdfoutput=1
\documentclass{article}

\usepackage{hyperref}       % hyperlinks
\usepackage[nonatbib, preprint]{neurips_2022}

\usepackage[utf8]{inputenc} % allow utf-8 input
\usepackage[T1]{fontenc}    % use 8-bit T1 fonts

\usepackage{url}            % simple URL typesetting
\usepackage{booktabs}       % professional-quality tables
\usepackage{amsfonts}       % blackboard math symbols
\usepackage{nicefrac}       % compact symbols for 1/2, etc.
\usepackage{microtype}      % microtypography
\usepackage{xcolor}         % colors
\usepackage{amsmath}
\usepackage{graphicx}
\usepackage{subfigure}
\usepackage{amssymb}
\usepackage{mathtools}
\usepackage{mathrsfs}
\usepackage{float}
\usepackage{multirow}
\usepackage{wrapfig}

\title{Teaching Networks to Solve Optimization Problems}

\author{% 
Xinran Liu$^{1}$ \quad
Yuzhe Lu$^{1}$\quad
Ali Abbasi$^{1}$\quad
Meiyi Li$^{2}$\quad
Javad Mohammadi$^{2}$\quad
Soheil Kolouri$^{1}$\\
$^1$Vanderbilt University \quad $^2$University of Texas at Austin\\
\texttt{\{xinran.liu, yuzhe.lu, ali.abbasi, soheil.kolouri\}@vanderbilt.edu}\\
\texttt{\{meiyil, javadm\}@utexas.edu}\\
}

\newcommand{\LOOP}{$\mathcal{LOOP}$}
\DeclareMathOperator*{\argmax}{arg\,max}
\DeclareMathOperator*{\argmin}{arg\,min}

\newcommand{\Xset}{\mathcal{X}}

\newcommand{\Uset}{\mathcal{U}}

\newcommand{\R}{\mathbb{R}}
\newcommand{\E}{\mathbb{E}}

\begin{document}
\maketitle

\begin{abstract}
 Leveraging machine learning to facilitate the optimization process is an emerging field that holds the promise to bypass the fundamental computational bottleneck caused by classic iterative solvers in critical applications requiring near-real-time optimization. The majority of existing approaches focus on learning data-driven optimizers that lead to fewer iterations in solving an optimization. In this paper, we take a different approach and propose to replace the iterative solvers altogether with a trainable parametric set function, that outputs the optimal arguments/parameters of an optimization problem in a single feed forward. We denote our method as Learning to Optimize the Optimization Process (\LOOP). We show the feasibility of learning such parametric (set) functions to solve various classic optimization problems including linear/nonlinear regression, principal component analysis, transport-based coreset, and quadratic programming in supply management applications. In addition, we propose two alternative approaches for learning such parametric functions, with and without a solver in the \LOOP.  Finally, through various numerical experiments, we show that the trained solvers could be orders of magnitude faster than the classic iterative solvers while providing near optimal solutions.
\end{abstract}

\section{Introduction}
\label{sec:intro}

Optimization problems are ubiquitous in computational sciences and engineering. Classic solutions to optimization problems involve iterative algorithms often relying on predetermined first and second order methods like (sub)gradient ascent/descent, conjugate gradients, simplex basis update, among others. These methods often come with desirable theoretical convergence guarantees, but their iterative nature could be limiting in applications requiring near-real time inference. Moreover, these algorithms' performance remains the same regardless of the number of times a similar optimization problem is visited. Recently, there has been an emerging interest in leveraging machine learning to enhance the efficiency of optimization processes and address some of these shortcomings. The learning based solutions are often referred to as \emph{Learning to Optimize} (L2O) methods in the literature. 

While L2O methods do not come with theoretical guarantees, they hold the promise of: 1) reducing the number of iterations needed to arrive at a solution, and 2) improving over time as more optimization problems are visited. L2O allows for transferring recent advances in machine learning, e.g., self-supervised learning, meta-learning, and continual learning, to learn data-driven optimization algorithms that could improve over time. Most existing L2O methods aim to learn a function that receives the current loss or its gradient, and based on the memory of previous loss values (or gradients) provide an update for the optimization parameters. Hence, these methods do not eliminate the iterative nature of the solution but aim at improving the iterative solution to: 1) reduce the number of total iterations, and 2) leading to better solutions for non-convex problems. 

% \begin{figure}[t!]
%     \centering
%     \includegraphics[width=0.7\textwidth]{Figures/LOOP_Concept.png}
%     \caption{We propose to replace the classic iterative solvers with a parametric (set) function that can be trained to directly map the input to the optimal parameters.}
%     \label{fig:concept}
% \end{figure}

    In this paper, we consider an inherently different use-case of machine learning in solving optimization problems. We propose to replace the classic iterative solutions of an optimization problem with a trainable parametric (set) function that directly maps the input of the optimization problem to the optimal parameters in a single feed forward. %Figure \ref{fig:concept} demonstrates this concept. 
    This process, which we denote as \emph{Learning to Optimize the Optimization Process} (\LOOP), is inspired by biological systems that are capable of solving complex optimization problems upon encountering the problem multiple times. By omitting the classic iterative solutions, \LOOP~overcomes one of the major optimization bottlenecks enabling near-real-time optimization in a wide range of critical applications. 

\LOOP~ is particularly suitable when one needs to perform a certain type of optimization (e.g., linear/quadratic programming) over a specific distribution of input data (e.g., sensors data collection) repeatedly. These problems abound in practice, with examples being cyber-physical infrastructures, autonomous vehicle networks, sensor networks monitoring a physical field, financial markets, and supply chains. For example, the resiliency and cost-effectiveness of our cyber-physical energy system relies on finding optimal energy dispatch decisions in near-real-time. This is a prime example of an optimization required to be repeatedly solved over the distribution of electricity demands on the power grid. Another example is traffic flow management in transportation networks, where traffic control systems need to determine traffic lights' status based on the traffic measurements continuously. 

At a first glance, the use of neural networks for solving frequently solved optimization problems may seem inefficient. However, such paradigm shift would allow us to leverage recent advances in deep learning, in particular, deep learning on edge-devices, continual learning, and transfer learning to improve the performance of an optimizer over time, even for a fixed given computational budget. Below we enumerate our specific contributions.
\begin{enumerate}
    \item Providing a generic framework, \LOOP, for replacing the classic iterative optimization algorithms with a trainable parametric (set) function that outputs the optimal arguments/parameters in a single feed forward.
    \item Proposing two generic approaches for training parametric (set) functions to solve a certain type of optimization problem over a distribution of input data.
    \item Demonstrating the success of our \LOOP~ framework in solving various types of optimization problems including linear/nonlinear regression, principal component analysis, the optimal transport-based coreset, and the quadratic programming in supply management application.
\end{enumerate}

\section{Prior Work}

One of the classic applications of machine learning in optimization has been in predicting proper hyper-parameters to solve an optimization problem. Such hyper-parameters could include learning rate, momentum decay, and regularization coefficients, etc. The existing literature on learning to predict hyper-parameters include approaches based on sequential model-based Bayesian optimization (SMBO)\cite{hutter2011sequential,bergstra2011algorithms,snoek2012practical}, and gradient-based methods \cite{bengio2000gradient,maclaurin2015gradient,wei2021meta}. At their core, these methods instantiate different variations of the same optimization algorithm, e.g., stochastic gradient descent (SGD), by selecting different hyper-parameters.
 
More recently, a large body of work has focused on leveraging machine learning to improve the optimization process by replacing the engineered optimizers with learnable ones. These methods, referred to as Learning to Optimize (L2O) approaches, are based on learning a parametric function, often in the form of a recurrent neural network, that receives the current loss (or its gradient) as input and outputs the parameter updates  \cite{gregor2010learning,li2016learning,andrychowicz2016learning,wichrowska2017learned,chen2017learning}. Such methods are effective in optimizing a wide range of optimization problems by reducing the number of iterations and often achieve better solutions for non-convex optimization problems. Chen et al. \cite{chen2021learning} provide a comprehensive review of these approaches and their numerous applications. Unlike the hyper-parameter search methods that instantiate different variations of the same optimization algorithm (e.g., SGD), L2O approaches effectively search over an expansive space of optimization algorithms to find an optimal algorithm. The optimal algorithm (i.e., the learned optimizer) fits input data distribution for a specific optimization problem (e.g., linear/quadratic programming); hence, it can lead to better performance than generic algorithms.

In this paper, our focus is entirely different from both hyper-parameter optimization approaches, and \emph{L2O} approaches discussed above. 
Instead of searching in the space of possible optimizers, our goal is to replace the optimization algorithm with a parametric (set) function that directly maps the optimization's input data to the optimal arguments/parameters. The motivation behind such transition is to: 1) discard iterations altogether, 2) have an optimizer that improves over time and encounters more optimization problems of a specific type. More importantly, the proposed framework allows one to leverage some of the core machine learning concepts, including continual/lifelong learning, transfer learning, domain adaptation, few/one/zero-shot learning, model compression (through sparse training and/or training), and many others into the improving the optimization process.

Several recent papers in the literature leverage deep neural networks to approximate the output of an optimization algorithm, which is in essence similar to our proposed framework, \LOOP. In VoxelMorph, for instance,  Balakrishnan et al. \cite{balakrishnan2019voxelmorph} trained a convolutional neural network to register medical images; image registration is a non-convex optimization problem often solved through time-consuming iterative and multi-scale solvers. 
In an entirely different application, Pan et al. \cite{pan2020deepopf} trained a neural network to predict the set of independent operating
variables (e.g., energy dispatch decisions) for optimal power flow (OPF) optimization problems, denoted as DeepOPF. They showed that DeepOPF requires a fraction of the time used by conventional solvers while resulting in competitive performance. 
More recently, Knyazev et al. \cite{knyazev2021parameter} trained a neural network to directly predict the parameters of an input network (with unseen architecture) to solve the CIFAR-10 and ImageNet datasets. \LOOP~ is the common theme behind these seemingly unrelated works. In this paper, we establish \LOOP~ as a generic alternative framework to the classic optimization algorithms, as well as, the L2O approaches, and show that many optimization problems can be directly solved through training neural networks.

 \begin{figure*}[t!]
    \centering
    \includegraphics[width=\linewidth]{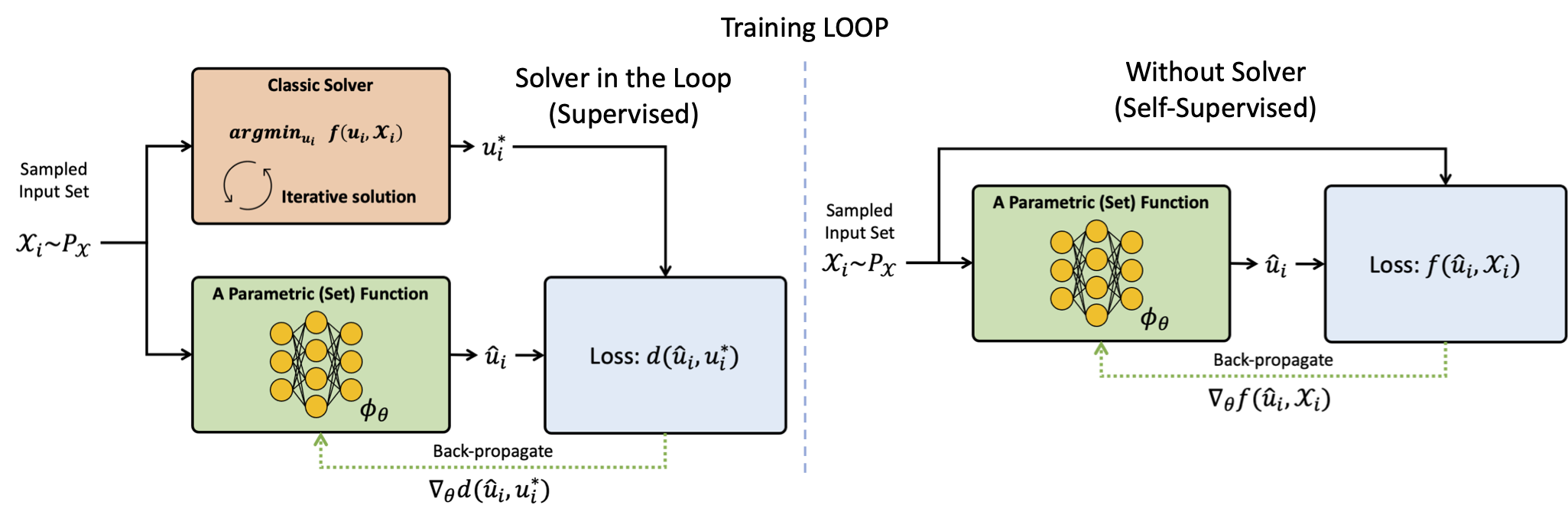}\vspace{-.5cm}
    \caption{Our two proposed approaches for training \LOOP: 1) with solver in the loop (left), and 2) without solver in the loop and by directly minimizing the objective function (right).}
    \label{fig:training}
\end{figure*}

% \subsection{Notations}
% Here we fix our notations for the rest of the paper. We will use $\R$ to identify real numbers, $\R^d$ to indicate the d-dimensional Euclidean space, bold small letters to indicate vectors, e.g., $\u\in\R^d$, bold capital letters to indicate matrices, e.g., $\X=[\x_1,...,\x_N]\in \R^{d\times N}$, sets are represented with caligraphic font, e.g., $\mathcal{X}$, where $|\mathcal{X}|$ denotes the set cardinality, functions are represented with small letters, e.g., $f(\cdot)$. We refer to the i'th element of a vector via $[\u]_i$, the (i,j)'th element of a matrix via $[\X]_{ij}$, and the $i$'th element of a set as $\{\Xset\}_i$.

\section{Method}
\label{sec:method}

We start by considering unconstrained optimization problems of the following type: 
\begin{align}
    u^*=\argmin_u f(\Xset,u)
\end{align}
where $\Xset=\{x_n\in \R^d \}_{n=1}^{N}$ is the set of inputs to the distribution, $u\in\mathbb{R}^l$ is the optimization parameters, and $f(\Xset,u)$ is the objective function with respect to parameters $u$ and inputs $\Xset$. To replace this optimization with a set function approximator, we propose two approaches as in Figure \ref{fig:training}.

{\bf Solver in the \LOOP--} In our first formulation, during the training, we use the classic solvers to obtain $u^*$ and use it as the ground truth. Then we pose the problem as a supervised learning problem. Our training objective is shown below:
\begin{align}
    \argmin_\theta~~& \E_{\Xset\sim P_\Xset}[ d(\phi_\theta(\Xset),u^*)]\nonumber ~~~~~~
    s.t.~~~ u^* = \argmin_u f(\Xset,u)
\end{align}
where $d(\cdot,\cdot):\R^l\times\R^l\rightarrow \R_+$ is a discrepancy/distance defined in $\R^l$, and $\phi_\theta$ denotes our set neural network, and $P_\Xset$ is a set distribution. 

{\bf Without Solver--} The use of a solver in our first formulation could be limiting, as such solvers are often computationally expensive turning the training excruciatingly slow. More importantly, in non-convex problems the calculated $u^*$ for input $\Xset$ is not unique (e.g., due to different initialization), which leads to solving a regression problem with changing targets. To avoid these problems, in our second formulation, we directly optimize the objective function and with a slight abuse of the term call it a ``self-supervised'' formulation:
\begin{align}
    \argmin_\theta~~& \E_{\Xset\sim P_\Xset}[ f(\Xset,\phi_\theta(\Xset))]
\end{align}
where the expected objective value over the distribution of the input sets is minimized. 
Note, for constrained problems (depending on the use case) we leverage different optimization techniques. For instance, we can enforce simple constraints (e.g., $u\geq 0$) into our model (i.e., the set function) using Rectified Linear Unit (ReLU) activations in the output layer of our network. Also, we can use the Lagrange dual function and absorb the constraints into our objective function as penalty terms. Next we describe the different optimization problems we consider in this paper. 

\subsection{Problem 1: Linear/Nonlinear Regression}
We start by the simple and yet routine problem of regression. Let $\Xset_i=\{(x_n^i\in \R^d, y_n^i\in\R)\}_{n=1}^{N_i}$ where the goal is to learn a parametric function $\rho_u:\R^d\rightarrow \R$. Here, index $i$ refers to the i'th regression problem of interest. In linear regression, $\rho_u(x)=u^T x$ (we absorbed the bias into $x$ for simplicity of notation). For nonlinear regression $\rho_u(x)=u^T\psi(x)$, $\psi:\R^d \rightarrow \R^l$ is a nonlinear mapping to a feature space (i.e., the kernel space). The optimization problem is then as follows:
\begin{align}
    u^*=\argmin_{u} \frac{1}{2}\sum_{n=1}^{N} \|\rho_u(x_n)-y_n\|_2^2+\lambda \Omega(u)
\end{align}
where $\Omega(u)$ is the regularization term (e.g., $\ell_2$ or $\ell_1$ norm), and $\lambda$ is the regularization coefficient. Our goal is then to learn a network that can solve the regression problem for unseen input data. 

% \subsection{Problem 2: Dictionary Learning} 

\subsection{Problem 2: Principal Component Analysis} 

Next, we consider the principle components analysis (PCA) problem, a common technique to project high-dimensional samples into a lower dimensional space while maximizing the variation of the data. Let $\Xset_i=\{x_n^i\in\R^d\}_{n=1}^{N_i}$, then PCA seeks an orthornormal set of $k$ vectors, $\{w_l\}_{l=1}^k$ such that:
\begin{align}
    w_l=\argmax_w&~~ w^T S_i w ~~~~~~ \nonumber
    s.t.~~w^T_jw_l=\left\{
    \begin{array}{lr}
        1 & j=l  \\
        0 & j<l
    \end{array}
    \right.
\end{align}
where $S_i=\frac{1}{N_i}\sum_{n=1}^{N_i}(x_n^i-\bar{x}^i)(x_n^i-\bar{x}^i)^T$ is the covariance matrix of the data, and $\bar{x}^i=\frac{1}{N_i}\sum_{n=1}^{N_i} x_n^i$ is the mean. 
Deriving the closed-form-solution for this problem
%This convex optimization problem leads to a closed form solution, which 
involves calculation of the eigenvectors of the covariance matrix, i.e., $S_iw^*_l=\lambda_l w^*_l$. 
% \begin{align*}
%     S_iw^*_l=\lambda_l w^*_l
% \end{align*}
Here $\lambda_l$ and $w^*_l$ are the l'th eigenvalue and eigenvector, respectively. This optimization problem can be presented as a set-function that receives a set of d-dimensional points, $\Xset^i$ with cardinality $|\Xset_i|=N_i$, and returns $U^*=[w_1^*,w_2^*,...,w_k^*]$. Using this representation, \LOOP  ~approximates the discussed set-function and outputs the top $k$ principle components for the input set. Hence, we aim to find a $\phi_\theta$, such that $\phi_\theta(\Xset)\approx U^*$ for $\Xset\sim P_\Xset$. 

\subsection{Problem 3: Optimal transport-based Coreset}
\begin{figure}
\vspace{-.8cm}
  \begin{center}
    \includegraphics[width=0.76\textwidth]{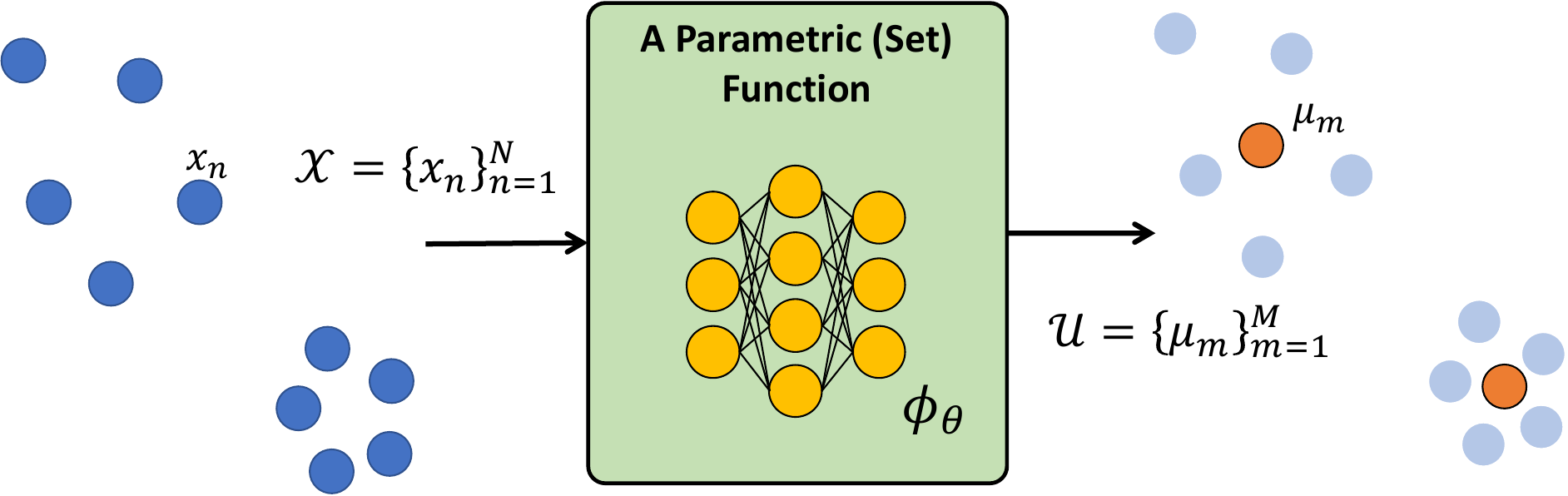}
  \end{center}
  \caption{For an input set $\Xset=\{x_n\in\mathbb{R}^d\}_{n=1}^N$, \LOOP~ returns a coreset $\Uset=\{\mu_m\in\R^d\}_{m=1}^M$ that minimizes the Wasserstein distance between the empirical distributions $p(x)=\frac{1}{N}\sum_{n=1}^N\delta(x-x_n)$ and $q_\Uset(x)=\frac{1}{M}\sum_{m=1}^M\delta(x-\mu_m)$}
  \label{fig:coreset}
\end{figure}
For our third problem, we consider the optimal transport-based coreset problem. The notion of coreset originates from computational geometry \cite{agarwal2005geometric} and has been widely used in machine learning tasks. Constructing a coreset from a large dataset is an optimization problem of finding a smaller set to best approximate the original dataset on a certain measure. Claici et al. \cite{claici2018wasserstein} leveraged optimal transport theory and introduced Wasserstein measure to calculate the coreset. 
Their work aims to minimize the Wasserstein distance of the coreset from a given input data distribution. In this paper we consider this transport-based coreset problem with respect to a fixed size output.

%For practical reasons, we  when the input distribution and the output coreset are finitely supported empirical distributions, and fixing the output size.

Let $\Xset=\{x_n\in \R^d\}_{n=1}^N$ be an input set. We assume that elements of each set are i.i.d. samples from an underlying distribution. Our sets are represented as empirical distributions, i.e., $p(x)=\frac{1}{N}\sum_{n=1}^N\delta(x-x_n)$. Given a size $M$ ($M\ll N$), we seek a set $\Uset^*=\{\mu_m\in\R^d\}_{m=1}^M$ with the empirical distribution $q_\Uset(x)=\frac{1}{M}\sum_{m=1}^M\delta(x-\mu_m)$, such that
\begin{align}
\Uset^*=\argmin_{\Uset} W_2(p,q_\Uset)
\end{align}
where $W_2(\cdot, \cdot)$ denotes the 2-Wasserstein distance. Existing approaches to this optimization problem rely on iterative linear programming to compute optimal transports in each iteration. We replace this costly process with a parametric set function $\phi_\theta$ such that $\phi_\theta(\Xset)\approx \Uset^*$ for $\Xset\sim P_\Xset$ (Figure \ref{fig:coreset}). The optimal transport-based coreset problem is equivalent to the free-support Wasserstein barycenter problem \cite{cuturi2014fast} when there is only one input distribution.

% \begin{figure}[H]
%     \centering
%     \includegraphics[width=0.6\textwidth]{Figures/LOOP_Coreset.pdf}
%     \caption{For an input set of $\Xset=\{x_n\in\mathbb{R}^d\}_{n=1}^N$, \LOOP~ returns a coreset $\Uset=\{\mu_m\in\R^d\}_{m=1}^M$ such that it minimizes the Wasserstein distance between the empirical distributions $p(x)=\frac{1}{N}\sum_{n=1}^N\delta(x-x_n)$ and $q_\Uset(x)=\frac{1}{M}\sum_{m=1}^M\delta(x-\mu_m)$, i.e.,  $W_2(p,q_\Uset)$.}
%     \label{fig:coreset}
% \end{figure}

\subsection{Problem 4: Supply management in Cyber-Physical Systems}
Lastly, we utilize \LOOP~to solve the fundamental problem of supply management in Cyber-Physical Systems (CPS). The electric power grid is an example of a CPS that is increasingly facing supply-demand issues. Power networks are large-scale systems spanning multiple cities, states, countries, and even continents and are characterized as a complex interconnect of multiple entities with diverse functionalities. The grid of future will differ from the current system by the increased integration of decentralized generation,
distributed storage, and communications and sensing technologies. These advancements, combined with climate change concerns, resiliency needs, and electrification trends, are resulting in a more distributed and interconnected grid, requiring decisions to be made at scale and in a limited time window. At its basic form, energy supply-demand problem seeks to find the most cost effective power production for meeting the end-users' needs and can be formulated as,
\begin{align}
\label{eq:ED}
    &\argmin_u \sum_{n=1}^{N} C_n(u_n) ~~~~~~~~s.t.~~~~ \sum_{n=1}^N u_n= \sum_{m=1}^M x_m,~\underline{u}_n \leq u_n \leq \overline{u}_n 
\end{align}

where $u_n$ is the produced electric power from source $n$ and $C_n$ is its corresponding cost, which is a quadratic function.  Given that $u_n$ represents the power output, it is bounded by physical limitation of the resource $n$, i.e., $\overline{u}_n $ and $\underline{u}_n$. In this setup, $x_m$ refers to the hourly electric demand in node $m$ (where the term `node' identifies an end-user/consumer). Note, the values of $x_m$ are positive. The equality constraint ensures the supply-demand balance. In practice, this problem is solved on an hourly basis to serve the predicted electric demand for the next hour. We aim to approximate this process with a parametric set function, such that $\phi_\theta(\Xset)\approx U^*$ for $\Xset\sim P_\Xset$.

\section{Experiments}
In this section, we demonstrate the application of \LOOP~ on problems enumerated in Section \ref{sec:method} and compare it to classic solvers. Throughout this section, GT refers to the Ground Truth and Solver refers to the results obtained from using commercial solvers to solve optimization problems of interest. For each problem and for each model architecture, we repeat the training of our \LOOP~ models five times, and we test the performance on a set of 100 problems per model. We then report the mean and standard deviations of all experiments over the five models and the 100 test sets.  We start by laying out the specifics of our models and then discuss the implementation details for each problem.

\subsection{Models}
Given that the inputs to our optimization problems are all sets, we pose these problems as learning permutation invariant deep neural networks on set-structured data. To that end, we use Deep Sets \cite{zaheer2017deep} with different pooling mechanisms and the Set Transformer \cite{lee2019set}.

{\bf Deep Sets} are permutation invariant neural architectures (i.e., the output remains unchanged under any permutation of input set's elements), which consist of: 1) a multi-layer perceptron (MLP) encoder, 2) a global pooling mechanism (e.g., average pooling), and 3) a MLP decoder that projects the pooling representation to the output; $\phi(\Xset)=\psi(pool(\{\eta(x_1),\cdots,\eta(x_n)\}))$.
    % \begin{align}
    %     \label{eq:DeepSets}
    %     \phi(\Xset)=\psi(pool(\{\eta(x_1),\cdots,\eta(x_n)\}))
    % \end{align}
    Here, $\eta$ is the encoder that extracts features from each element of $\Xset$ independently, resulting in a permutation equivariant function on the input set, and $\psi$ is the decoder that generates final output after a pooling layer ($pool$). To achieve a permutation invariance set function, the pooling mechanisms must be a permutation invariance operator (e.g., average pooling, or more advanced methods like Pooling by sliced-wasserstein embedding (PSWE) \cite{naderializadeh2021pooling}). Specifically, we use global average pooling (GAP) and Sliced-Wasserstein Embedding (SWE) \cite{naderializadeh2021pooling,lu2021slosh} respectively as the pooling layer. 
    
    {\bf Set Transformer} follows a similar blueprint of permutation equivariant encoder, permutation invariant pooling, and permutation equivariant decoder as Deep Sets. However, while the encoder in the Deep Sets model acts on each set element independently, Set Transformers use attention to pass information between elements in the encoder. This allows the encoder to model relations between elements, which can be crucial to approximate a parametric (set) function in some learning tasks.  
    
    More precisely, the encoder is a stack of multiple trainable (Induced) Set Attention Blocks (SAB and ISAB) \cite{lee2019set} that perform self-attention operations on a set and produce output containing information about pairwise relations between elements. Note that these blocks are permutation equivariant, that is, for any permutation $\pi$ of elements in $\Xset=\{x_i\}_{i=1}^n$, $block(\pi\Xset)=\pi block(\Xset)$. As a composition of permutation equivariant blocks, the encoder is also permutation equivariant and captures higher-order interaction features. The decoder aggregates features by a learnable pooling layer, Pooling by Multihead Attention (PMA) \cite{lee2019set}, and send them through a SAB to get output. Since PMA is a permutation invariant operator, and the rest of the operators (SAB or ISAB) are all permutation equivariant, Set Transformer becomes a permutation invariant architecture.

\begin{figure}[t!]
    \centering
    \includegraphics[width=\linewidth]{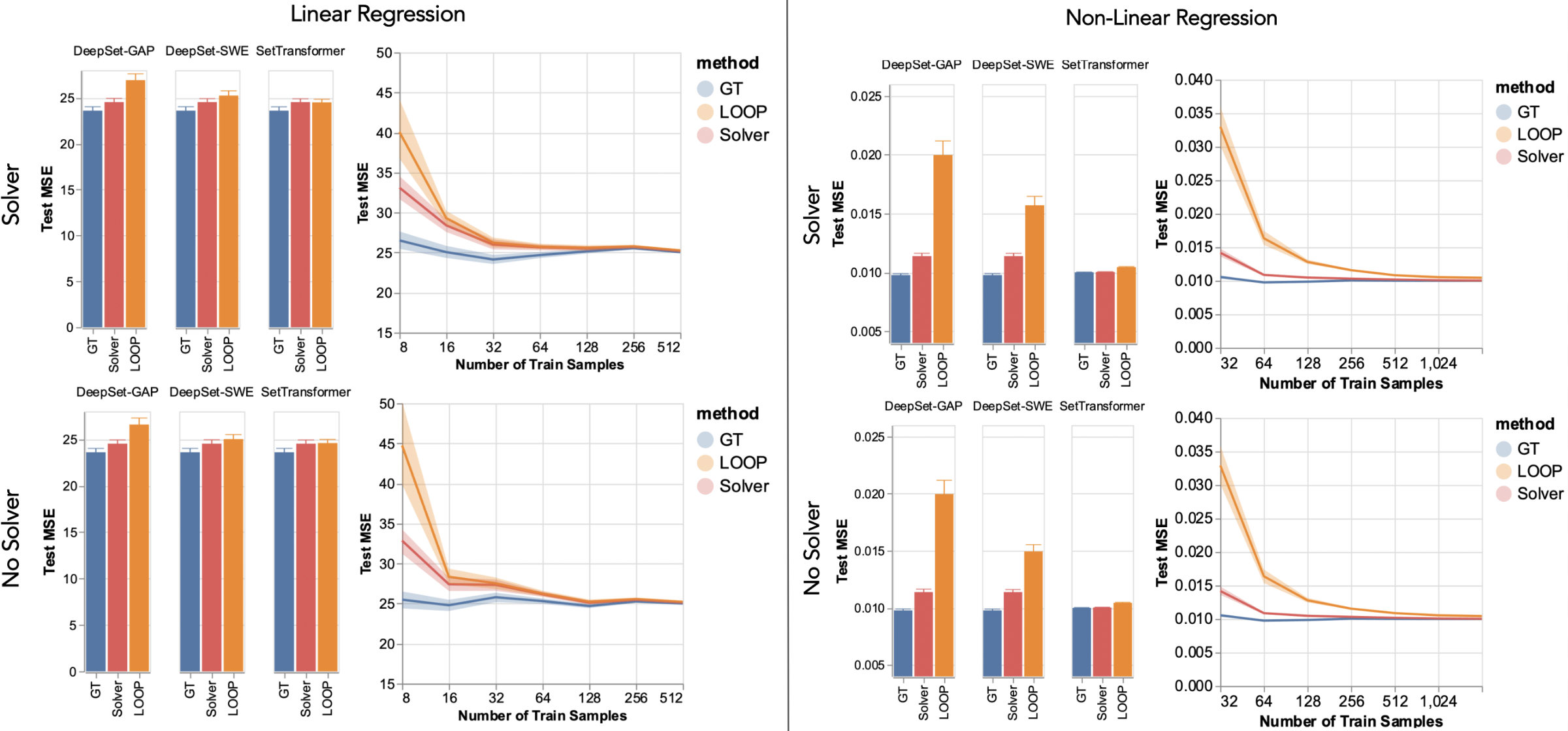}\vspace{-.2in}
    \caption{Performance comparison between \LOOP~and the solver for three different model architectures (left) and for the two proposed learning settings (with or without the solver in the \LOOP) for the linear regression (a) and nonlinear regression (b) problems. The plots on the right shows the performance of the Set Transformer network and the solver as a function of training samples.}
    \vspace{-.3cm}
    \label{fig:regression}
\end{figure}

\subsection{Problem 1: Linear/Nonlinear Regression}
\label{subsec:reg}
\textit{Dataset}: We follow a generative model $y=w^T\phi(x)+\epsilon$, where $\phi(\cdot)$ is the feature map, $w$ contains the ground truth parameters of our regression problem, and $\epsilon$ denotes noise. For feature maps, in the linear case we have $\phi(x)=[1,x]^T$ and in the nonlinear case, we select $\phi(x)=[\rho(x-\mu_1),...,\rho(x-\mu_M)]$ with $\rho(x)$ being a radial basis function and $\{\mu_m\}_{m=1}^M$ form a grid in a predefined interval. %(e.g., $[-10,10]$). 
To generate each dataset  $\Xset_i$, we first sample the set cardinality $N_i$ uniformly from a predefined interval.  Then, we  sample $w$, $\{\epsilon_n\}_{n=1}^{N_i}$, and $\{x_n\}_{n=1}^{N_i}$, and generate our $(x^i_n,y^i_n)$ pairs (train and test). 

% first choose the ground truth parameters $a$ and $b$ randomly to generate data points on the line $y=ax+b$ in 2D. Then we add Gaussian noise to the data and solve the linear regression problem on this dataset. The dataset for the nonlinear regression problem is generated using Gaussian basis functions. At randomly selected points in a specific range, we get values of a linear combination of the basis functions with random weights, and add a Gaussian noise to the dataset.

\begin{figure}
    \centering
    \includegraphics[width=\linewidth]{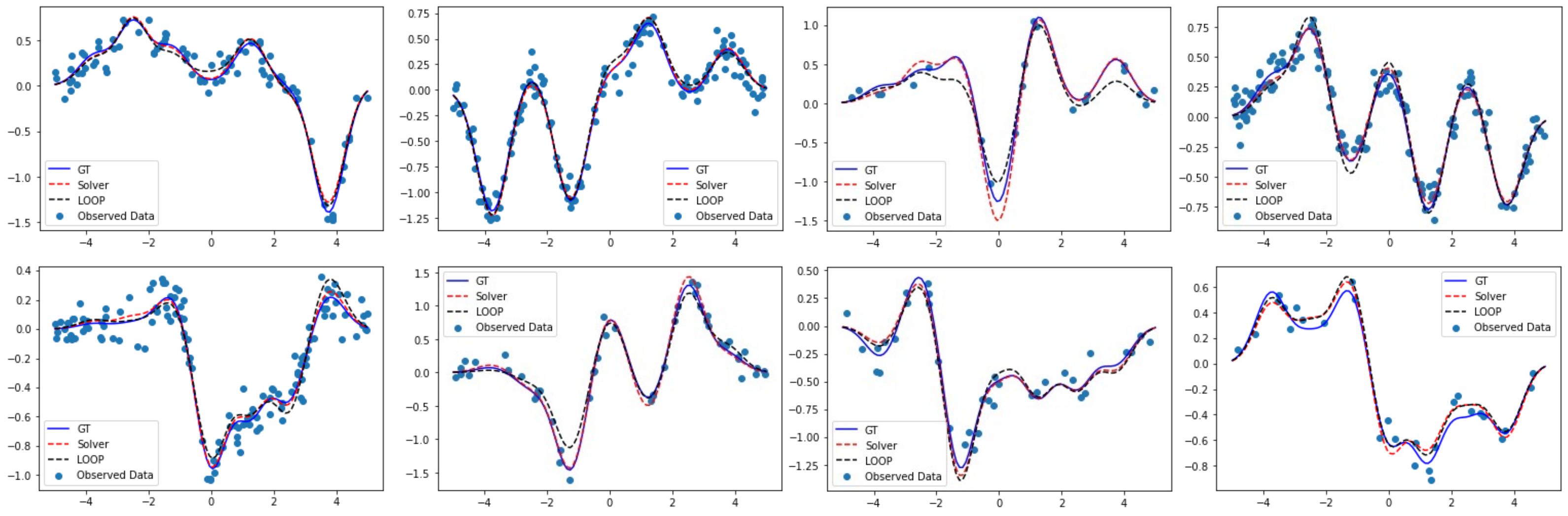}
    \caption{Sample qualitative results for our nonlinear regression problem compared with the Ground Truth (GT) and the solver's result.  Note that the input set cardinality is a random variable and it varies among these plots. }
    \label{fig:qual}
\end{figure}

For each model architecture and each learning setting (i.e., with and without solver in the \LOOP) we train our \LOOP~ model 5 times and report the test MSE of our model, the solver, and the ground truth. Figure \ref{fig:qual} shows sample qualitative results of our nonlinear regression experiments with ground truth, solver, and \LOOP~results overlayed on the observed noisy data. In addition, for the Set Transformer architecture, we report the test performance of our trained \LOOP~ model and the solver as a function of the number of training samples (Figure \ref{fig:regression}). We see that while for all architectures \LOOP~ is able to perform comparable with the solver, for the Set Transformer architecture the gap between \LOOP~ and the solver is the smallest.

% \begin{figure}[t!]
%     \centering
%     \includegraphics[width=\columnwidth]{Figures/NonLinearReg.pdf}
%     \caption{Non-Linear regression}
%     \label{fig:nonlinearreg}
% \end{figure}

\begin{figure}[t!]
    \centering
    \includegraphics[width=\linewidth]{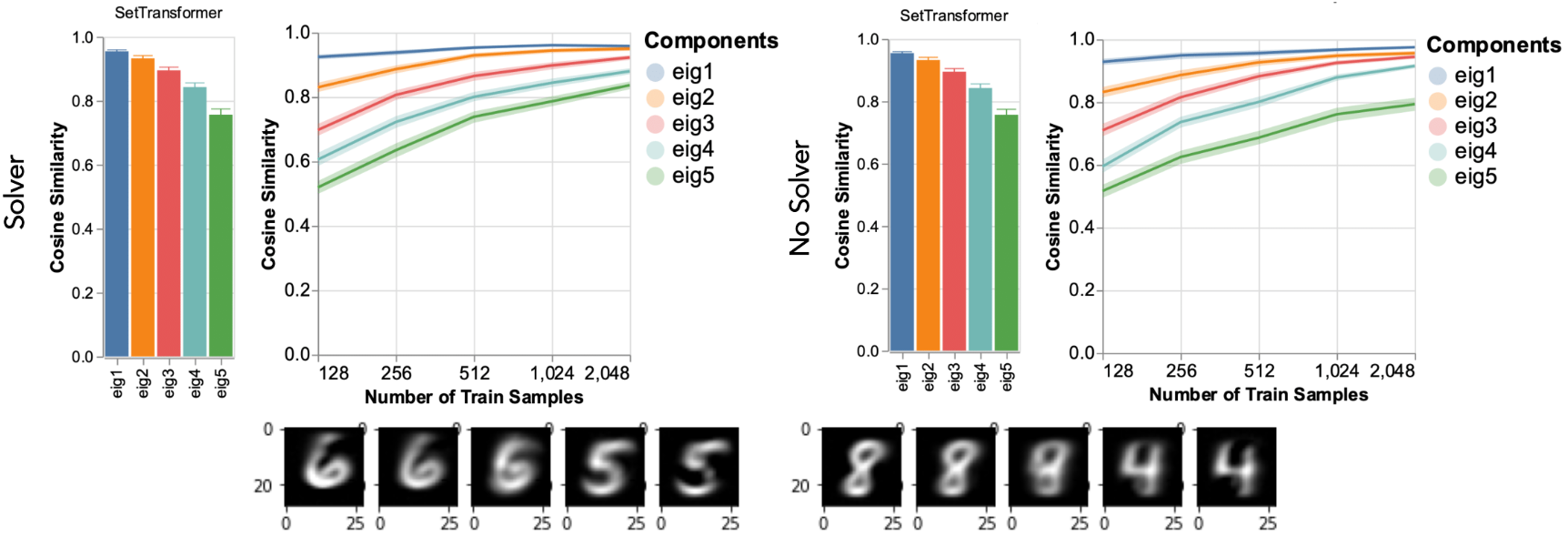}\vspace{-.2in}
    \caption{\LOOP's~ performance in predicting the principle components as measured by the cosine similarity between the solver's and our model's outputs. We also provide the performance of the network as a function of the input data cardinality. On the bottom is the visualization of the first eigenvector calculated by our \LOOP~model on four different problems with two random pairs of digits. We can see that the network's output is quantitatively and qualitatively aligned with the first principle component.}
    \label{fig:pca}
\end{figure}

% \begin{figure}[top!]
%     \centering
%     \includegraphics[width=\columnwidth]{Figures/PCA.pdf}
%     \caption{Qualitative result of PCA}
%     \label{fig:pca-qual}
% \end{figure}

\begin{figure}[t!]
    \centering
    \includegraphics[width=\columnwidth]{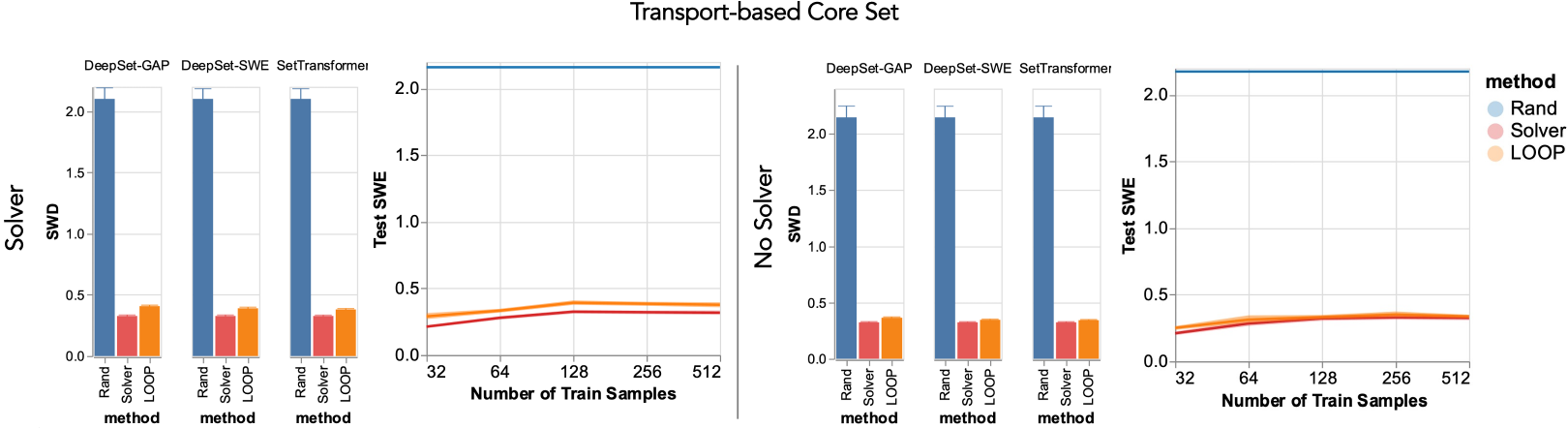}
    \caption{Performance comparison between \LOOP~and the solver for our three different models (left) under the two proposed learning settings. The y-axis represents the average Wasserstein distance between the input distribution $p$ and the coreset distribution $q_\Uset$, when the coreset are: 1) random samples from the uniform distribution, 2) the output of the solver, and 3) the output of our \LOOP~model. The plots on the right of each column show the performance of the Set Transformer and the solver as a function of the number of training samples.}
    \label{fig:barycenter}
\end{figure}

\subsection{Problem 2: Principal Component Analysis}

 \textit{Dataset}: We used the MNIST \cite{lecun1998mnist} dataset to sample train and test sets. MNIST contains 60,000 train and 10,000 test images of handwritten digits. The size of a single image is $28\times 28$. During training, we first select pairs of random digits to sample images from. Then a random number of data ranging from 500 to 1000 is uniformly sampled from the two digits to form the input set.
 
 Given an input set $\Xset_i$, our network aims to predict the top $K=5$ eigenvectors of the input data. In ``solver in the \LOOP" approach, the top $K=5$ eigenvectors are calculated by the solver. Then our set transformer \cite{lee2019set} is trained to maximize the cosine similarities between the ground-truth eigenvectors and the predicted ones. In our ``no solver" approach, the set transformer maximizes the area under the curve of the captured variances along the predicted eigenvectors. We train 5 different models for this experiment and evaluated each model on 100 different test problems. For our metric, we calculate the cosine similarities between the predicted vectors and the principle components obtained from the solver. The mean and standard deviation of the cosine similarities for each eigenvector is depicted in Figure \ref{fig:pca} (left). We also show the performance of the trained model as a function of different number of training samples from 128 to 2048 (on the right). Results of ``solver in the \LOOP" and ``no solver" training are shown for the Set Transformer model in \ref{fig:pca}. Our network is able to effectively predict the top principle components in all experiments, while having a higher fidelity for the ones with larger eigenvalues.  %and the cosine similarities drop while moving to less important principal components. 
% The reduction in similarities are acceptable considering the fact that usually the first principal component contains the most amount of variance information. This is qualitatively shown in \ref{fig:pca-qual}. It illustrates that as we are deviating from the mean of the data, by  Moreover, \ref{fig:pca} demonstrates as we are increasing the number of data, the set transformer can predict the principal components more accurately. 

\subsection{Problem 3:  Transport-based coreset}
\label{subsec:ot}
\textit{Dataset}: We generate datasets by drawing samples from random 2D Gaussian Mixture Models (GMMs). We start by initializing a random number of Gaussians with random means but a fixed covariance matrix. Then, we draw random number of samples from each of these randomly initialized Gaussians to generate our input sets $\Xset^i$.  

% In short, we first draw random samples from a random number of Gaussians, each with a random mean and a fixed covariance matrix and shuffle the data. 

\textit{Results:} Results of the two training approaches ``solver in the \LOOP" and ``no solver" using three different models are shown in \ref{fig:barycenter}. Since the problem is equivalent to the free-support Wasserstein barycenter problem, we used the solver from the Python Optimal Transport package \cite{flamary2021pot} as the solver. To compare the output of our \LOOP~model with the solver, we use $W_2(p,q_\Uset)$ as our metric (the lower the better). Also, to provide a reference for comparison, we also consider the Wasserstein distance between the input distribution, $p$, and a uniform distribution in the input domain, $\bar{q}$, which we refer to as Rand (equivalent to chance). We used the Sliced-Wasserstein distance (SWD) \cite{kolouri2019generalized} as the objective function in the ``no solver" training, as SWD is significantly faster to compute than the Wasserstein distance. Finally, we compare the performance of \LOOP~ and the solver as a function of number of training samples in Figure \ref{fig:barycenter}.

\subsection{Problem 4: Supply management in Cyber-Physical Systems}

\textit{Dataset:} %\textcolor{red}{Meiyi}
    We use the publicly available IEEE 2000-bus system data set \cite{xu2017application} as the seed infromation to generate hourly energy data for one week. We use different load profiles for weekdays and weekends and randomly scale the original data. The scaling coefficient lies between 0.95 and 1.05.  This process results in $24\times 7$ data points. We use the data of odd hours for training and that of even hours for testing. The IEEE 2000-bus system is a 2,000 nodes graph representing a realistic large scale electric grid. This network consists of 1,125  demand nodes and 544 supply nodes. %The number of these supply and demand nodes define the dimensions of the input and the output for our network.
    
\textit{Results:} 
For solver in the \LOOP, we use the mean squared error as the loss function, i.e., $L=\frac{\sum_{n=1}^{N}(u_n-u_n^{*} )^2   }{n}$.
% \begin{align}
%     & L=\frac{\sum_{n=1}^{N}(u_n-u_n^{*} )^2   }{n} \label{eq:ED_loss_n}
% \end{align}
Here $\{u_n^*\}_{n=1}^N$ are the solver's output. We use the quadratic programming (QP) solver of CVXPY library \cite{diamond2016cvxpy} as our solver. 
In our second learning setting, i.e., with no solver in the \LOOP, we include the optimization constraints in our objective as penalty terms. Therefore, the loss function will consist of three terms, 
{\small
\begin{align}
    &L=\sum_{n=1}^{N} C_n(u_n)+\lambda _1\left ( \sum_{n=1}^N u_n- \sum_{m=1}^M x_m \right )^2
    +\lambda _2\left [ (ReLU(\underline{u}_n-u_n))^2+ (ReLU(u_n-\overline{u}_n))^2\right ]\label{eq:ED_loss_n_2}
\end{align}}
where $\lambda_i$s are the penalty coefficients. We use $\lambda_1 =0.001$, $\lambda_2 =10$ in our experiments. 

\begin{figure}
\vspace{-.2cm}
  \centering
    \includegraphics[width=0.8\textwidth]{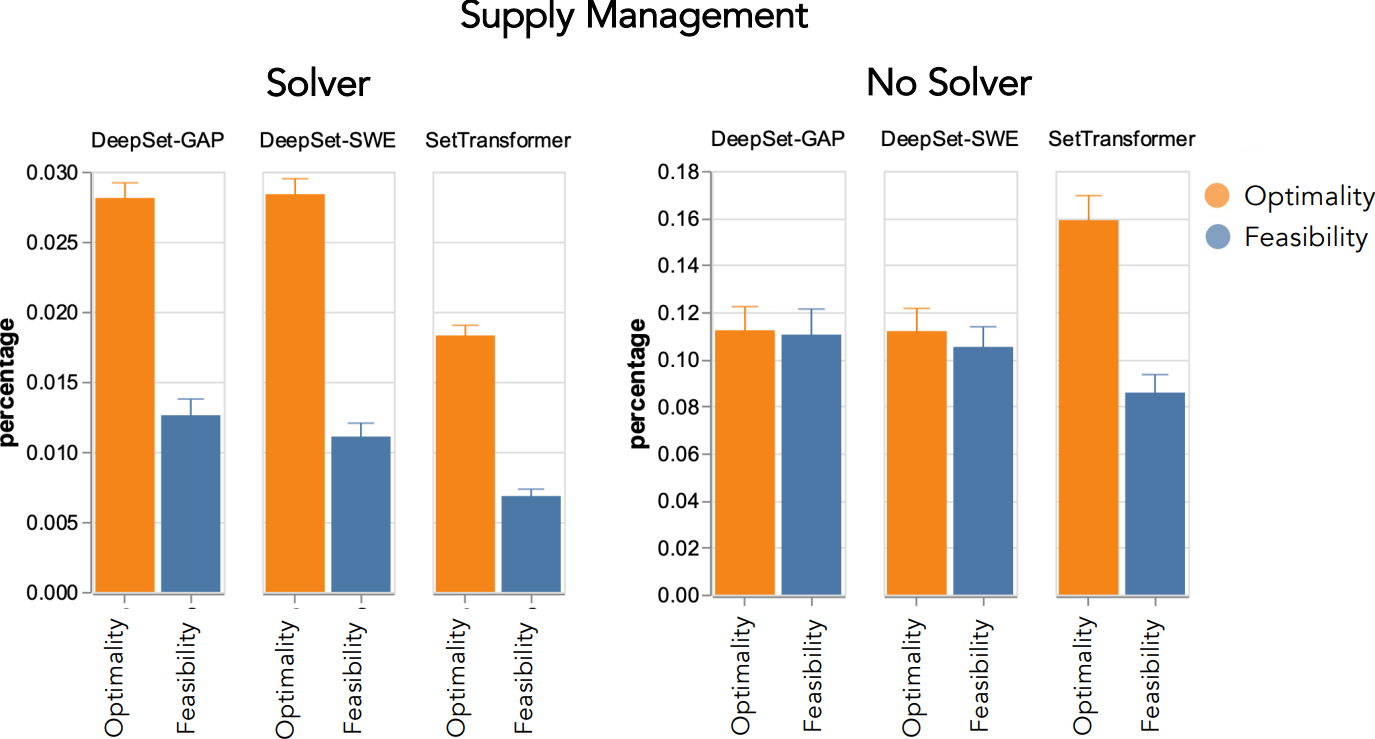}
  \caption{\LOOP~'s performance measured by the distance of its output from the optimal solution and the feasible set %(the set of solutions satisfying the constraints). 
  We define the optimality distance as $\sum_{n=1}^{N}\left | u_n-u_n^* \right | / \sum_{n=1}^{N}u_n^*$ , where $u_n^*$ and $u_n$ refer to the solver's and \LOOP's~outputs. The feasibility distance $\sum_{n=1}^{N}\left | u_n-u_n^{proj} \right | /  \sum_{n=1}^{N}u_n^{proj}$ where $u_n^{proj}$ denotes projection of $u_n$ onto the feasible set.}
  \label{fig:supply_management}
  \vspace{-.6cm}
\end{figure}

In \ref{eq:ED_loss_n_2}, the first term represents the cost of electricity production. The second term ensures the equality of supply and demand, and the third term enforces the the supply to be bounded. We bound the output according to inequality constraints for testing.
% \begin{figure}[t!]
%     \centering
%     \includegraphics[width=0.8\textwidth]{LOOP_NeurIPS/Figures/supply management.png}
%     \caption{Performance of our \LOOP~model as measured by the distance of its output from the optimal solution and from the feasible set (i.e., the set of solutions that satisfy the constraints). We define the optimality distance as $\sum_{n=1}^{N}\left | u_n-u_n^* \right | / \sum_{n=1}^{N}u_n^*$ , where $u_n^*$ and $u_n$ refer to the solver's and \LOOP's~outputs. The feasibility distance is derived by $\sum_{n=1}^{N}\left | u_n-u_n^{proj} \right | /  \sum_{n=1}^{N}u_n^{proj}$ where $u_n^{proj}$ denotes projection of $u_n$ onto the feasible set. This figure depicts performance for \LOOP~and the solver for our three different architectures, with and without solver.
%     }
%     \label{fig:supply_management}
% \end{figure}
To quantify the performance of our \LOOP~model, we report two metrics: 1) optimality, which measures how far we are from the solver's output, and 2) feasibility, which measures the distance of \LOOP's output from the feasible set. We measure feasibility distance by projecting the network's output onto the feasible set and measuring the distance between the original and project solutions.

Figure \ref{fig:supply_management} shows the results for two \LOOP~approaches (solver in the \LOOP and no solver) using different models. The gap between our two learning settings for this problem is more significant than previous unconstrained ones. 

This is because the \LOOP~ model with no solver minimizes a combination of the objective function and the penalty terms. Unlike the solver in the \LOOP~ model (which could leverage optimality information), the \LOOP~ model with no solver does not establish any relation between feasibility and optimality. We also present results of different penalty parameters %($\lambda_1=\lambda_2$) 
for the \LOOP~model with no solver in the supplemental materials, where the gap between the two \LOOP~approaches is reduced by more careful tuning of penalty terms ($\lambda_1$ and $\lambda_2$).  
% \begin{figure}[t!]
%     \centering
%     \includegraphics[width=0.8\columnwidth]{LOOP_NeurIPS/Figures/lambda.pdf}
%     \caption{Performance of \LOOP~on the supply management problem using different penalty parameters($\lambda_1=\lambda_2$) .}
%     \label{fig:lambda}
% \end{figure}
Moreover, Table \ref{T:time} presents the average computing time %of one data point 
of different solvers (ECOS \cite{6669541}, CVXOPT \cite{vandenberghe2010cvxopt}, OSQP \cite{stellato2020osqp}, Matpower 7.1 \cite{zimmerman2010matpower}), and \LOOP~over ten runs. Using GPU, three \LOOP~approaches outperform all solvers. On CPU, the \LOOP~model of Set Transformer performs on par with the today's solvers while other \LOOP~models are noticeably faster.

\begin{table}[t!]
\caption{Average computing time of %one data point 
using different solvers and \LOOP~approaches over ten runs.}
\label{T:time}
\centering
\begin{tabular}{|c|c|c|c|c|c|} 
\hline
                                                                   & \textbf{Method}     & \textbf{Time(s)} &                                                                           & \textbf{Method}      & \textbf{Time(s)}  \\ 
\hline
\multirow{7}{*}{\begin{tabular}[c]{@{}l@{}}\rotatebox[origin=c]{90}{Solvers}\\\end{tabular}} & ECOS Solver %\cite{6669541}
(CPU)    & 0.1237           & \multirow{6}{*}{\begin{tabular}[c]{@{}l@{}}\rotatebox[origin=c]{90}{ $\mathcal{LOOP}$}\end{tabular}} & SetTransformer(CPU)~ & 0.1284            \\ 
\cline{2-3}\cline{5-6}
                                                                   & CVXOPT Solver(CPU)  & 1.8277           &                                                                           & DeepSet-SWE(CPU)     & 0.0600            \\ 
\cline{2-3}\cline{5-6}
                                                                   & OSQP Solver(CPU)    & 0.1421           &                                                                           & DeepSet-GAP(CPU)     & 0.0538            \\ 
\cline{2-3}\cline{5-6}
                                                                  & ECOS Solver(GPU)    & 0.1005           &                                                                           & SetTransformer(GPU)~ & 0.0057            \\ 
\cline{2-3}\cline{5-6}
                                                             & CVXOPT Solver(GPU)  & 1.6288                   &                                                                           & DeepSet-SWE(GPU)     & 0.0070            \\ 
\cline{2-3}\cline{5-6}
                                                           & OSQP Solver(GPU)    & 0.1218                    &                                                                           & DeepSet-GAP(GPU)     & 0.0022            \\ 
\cline{2-6}
                                          & Matpower 7.1 Solver & 0.9609                                  &                                                                           &                      &                   \\
\hline
\end{tabular}
\vspace{-.2cm}
\end{table}
% \begin{align}
%     &\textup{optimality gap } =\frac{\sum_{n=1}^{N}\left | u_n-u_n^* \right | }{\sum_{n=1}^{N}u_n^*} \label{eq:ED_og}\\  
%     &\textup{feasibility gap }=\frac{\sum_{n=1}^{N}\left | u_n-u_n^{proj} \right | }{\sum_{n=1}^{N}u_n^{proj}} \label{eq:ED_fg}
% \end{align}

\subsection{Continual learning on \LOOP}

A promising capability of \LOOP~is to continually train the network to adjust to the change in the input distribution. Note, \LOOP~(specifically in the no solver in the \LOOP~ setting) is ripped for continual learning. In short, the \LOOP~agent can evaluate the quality of its prediction (i.e., by measuring the objective value or by checking the feasibility) and perform continual learning if the prediction quality is degraded. We perform a continual learning experiment on \LOOP~, while the agent is tasked to solve nonlinear regression problems where the frequency spectrum of the input data drifts from Task 1 to Task 2. To overcome catastrophic forgetting, we use memory replay \cite{van2020brain} as one of the core bio-inspired mechanisms for overcoming catastrophic forgetting \cite{kudithipudi2022biological}. Table \ref{T:cl} demonstrates the application of \LOOP~ in continual learning of non-linear regression under domain shift. We see that memory replay enables \LOOP~to learn the new task (Task 2) while achieving positive backward transfer on Task 1. More details are provided in the supplementary materials. Lastly, the application of other continual learning mechanisms like regularization-based approaches \cite{delange2021continual} and gradient projection approaches \cite{saha2020gradient,abbasi2022sparsity}, opens up an exciting research direction for future work.

\begin{table}[H]
    \caption{Test MSE for both tasks after each training phase using \LOOP~(left) and \LOOP~with memory replay (right). The models are trained on task 1 in phase 1 and on task 2 in phase 2.}
    % In phase 1, we train on the task 1 and in phase 2 we train on task 2. \LOOP~(left) suffers from catastrophic forgetting for task 1 after training on task 2, while \LOOP~with memory replay (right) has positive backward transfer on task 1 after learning task 2.}
    \begin{minipage}{0.5\linewidth}
      \centering
        \begin{tabular}{|c|c|c|}
        \hline
         & Task 1 & Task 2\\
        \hline
        test MSE after phase 1 &0.091 &4.805 \\
        \hline
        test MSE after phase 2 & 0.287& 0.130\\
        \hline
        \end{tabular}
    \end{minipage}%
    \begin{minipage}{0.5\linewidth}
      \centering
        \begin{tabular}{|c|c|c|}
        \hline
         & Task 1 & Task 2 \\
        \hline
        test MSE after phase 1 & 0.091 & 4.805\\
        \hline
        test MSE after phase 2 & 0.088 & 0.136\\
        \hline
        \end{tabular}
    \end{minipage} 
    \label{T:cl}
    \vspace{-.3cm}
\end{table}

\section{Conclusion}
This paper presents a novel alternative for existing iterative methods to solve optimization problems. Specifically, this paper introduces \LOOP~(Learning to Optimize Optimization Process) framework, which approximates the optimization process with a trainable parametric (set) function. Such a function maps optimization inputs to the optimal parameters in a single feed forward. We proposed two approaches for training \LOOP; using a classic solver for providing ground truth (supervised learning) and without a solver in the \LOOP~(self-supervised learning). The performance of the proposed methods is showcased in the contexts of diverse optimization problems; (i) linear and non-linear regression, (ii) principal component analysis, (iii) transport-based coreset, and (iv) supply management in cyber-physical setups. We used three separate models in our experiments, namely
deep sets with global average pooling,% (GAP), 
deep sets with sliced-Wasserstein Embedding, and Set Transformers. Our results supports that replacing optimization problem with a single forward mapping yields outputs within a reasonable distance from commercial solvers' solutions. \LOOP~holds the promise for the next generation of optimization algorithms that improve by solving more optimization problems.  
%In future, we intend to leverage recent advancements in deep learning on edge-devices, continual learning and transfer learning to continuously improve \LOOP's~ performance over time. 

% \section{Section Title}
% Main contents here.

% \subsection{Subsection Title}
% A figure in Fig.~\ref{fig:spiral}. Please use high quality graphics for your camera-ready submission -- if you can use a vector graphics format such as \texttt{.eps} or \texttt{.pdf}.
% \begin{figure}[htp]
% \begin{center}
% \includegraphics[width=0.5\textwidth]{spiral.eps}
% \caption{A spiral.}\label{fig:spiral}
% \end{center}
% \end{figure}

% An example of citation~\cite{DBLP:conf/acml/2009}.

%\acks{Acknowledgements should go at the end, before appendices and references. You can uncomment this for the camera-ready version on paper acceptance.}

\bibliographystyle{plain}
\bibliography{loop}

\end{document}